\def\year{2021}\relax
\documentclass[letterpaper]{article} 
\usepackage{aaai21}  
\usepackage{times}  
\usepackage{helvet} 
\usepackage{courier}  
\usepackage[hyphens]{url}  
\usepackage{graphicx} 
\usepackage{natbib}  
\usepackage{caption} 
\usepackage{amsfonts} 
\usepackage{booktabs}       
\usepackage{multirow}
\usepackage{xcolor}
\usepackage[caption=false,font=footnotesize]{subfig}
\usepackage{tikz}
\usetikzlibrary{decorations.pathmorphing}

\title{Learning Abstract Task Representations}
\author {
    Mikhail M. Meskhi,\textsuperscript{\rm 1}
    Adriano Rivolli,\textsuperscript{\rm 2}
    Rafael G. Mantovani,\textsuperscript{\rm 3}
    Ricardo Vilalta,\textsuperscript{\rm 1} 
    \\
}
\affiliations {
    \textsuperscript{\rm 1} Department of Computer Science, University of Houston, Houston, TX, 77204 \\
    \textsuperscript{\rm 2} Department of Computing, Universidade Tecnol\'ogica Federal do Paran\'a, Corn\'elio Proc\'opio, Brazil \\
    \textsuperscript{\rm 3}  Universidade Tecnol\'ogica Federal do Paran\'a, Apucarana, Brazil  \\
    \{mmeskhi, rvilalta\}@uh.edu, 
    \{rivolli, rafaelmantovani\}@utfpr.edu.br
}

\begin{document}

\maketitle


\begin{abstract}
A proper form of data characterization can guide the process of learning-algorithm selection and model-performance estimation. The field of meta-learning has provided a rich body of work describing effective forms of data characterization using different families of meta-features (statistical, model-based, information-theoretic, topological, etc.). In this paper, we start with the abundant set of existing meta-features and propose a method to induce new abstract meta-features as latent variables in a deep neural network. We discuss the pitfalls of using traditional meta-features directly and argue for the importance of learning high-level task properties. We demonstrate our methodology using a deep neural network as a feature extractor. We demonstrate that 1) induced meta-models mapping abstract meta-features to generalization metrics outperform other methods by $\sim 18\%$ on average, and 2) abstract meta-features attain high feature-relevance scores.
\end{abstract}



\section{Introduction}
\label{sec:intro}

Meta-learning (MtL) allows rational agents to improve on their learning abilities through a process known as \textit{learning to learn}~\cite{hospedales2020metalearning,schmidhuber1987evolutionary,vanschoren2018meta,  vilalta2002perspective}.  One major goal is to build self-adaptive systems that adjust their learning mechanism automatically with new tasks. Automatic adaptation can be described in a plethora of ways; it can be as simple as tuning hyper-parameters, selecting a different family of learning algorithms, or simply warm-starting a model. Meta-learning relies on past experience stored in the form of \textit{meta-knowledge}. One type of meta-knowledge encompasses families of meta-features used as a form of data (or task) characterization. Meta-features capture various types of data properties  such as statistical, e.g., number of numerical attributes; degree of class separation, e.g., Fisher's Linear Discriminant~\cite{Ho2002}; or level of concept complexity, e.g., concept variation~\cite{vilalta1999understanding,rendell_concept}. The proper identification of data properties is essential to map tasks to learning mechanisms. 

Several approaches in meta-learning use families of meta-features as input to quantify task similarity. It is common to compute task similarity as the Euclidean distance between two meta-feature vectors. While this approach has shown to be effective in simple scenarios~\cite{vanschoren2018meta}, it exhibits clear limitations. First, selecting a subset of relevant meta-features is a non-trivial task. What criteria should we invoke to select or discard a family of meta-features? For example, statistical meta-features are not always intuitive and lack expressiveness. Previous work has shown how different datasets may share identical statistical properties but markedly different data distributions~\cite{Matejka}. 
Second, computing certain types of meta-features on large datasets is computationally costly. For example, topological meta-features perform multiple passes over the training dataset to compute a single figure of merit~\cite{Ho2002, Lorena2018}. Ultimately, the selection of meta-features is an ad hoc process based on domain knowledge.

In this paper, we propose an approach that learns abstract meta-features from families of traditional meta-features using a deep neural network. 
We argue that traditional meta-features are not always capable of capturing crucial task characteristics. This can be attributed to inherent limitations such as being hand-crafted, and not being tuned to specific tasks (lacking in general applicability). Extracting meta-features on large datasets can quickly scale up computational costs and execution times. For example, complexity meta-features involve geometrical computations resulting in long computing times. 

The main contribution of this paper is a novel method of inducing abstract meta-features as latent variables learned by a deep neural network; experimental results demonstrate the efficacy of the learned abstract meta-features in improving generalization performance estimation.


\section{Related Work}
\label{sec:related}







Data characterization consists of extracting meaningful task properties. Simple, statistical, and information-theoretic meta-features can be straightforwardly extracted from datasets by capturing information concerning data dimensionality, distribution, and the amount of information present in the data. Model-based and landmarking meta-features characterize datasets indirectly by using the induced learning models; these meta-features comprise model properties and model performance~\cite{rivolli2018towards}. 

Another family of meta-features is based on capturing the complexity of a learning task~\cite{Ho2002}, and has been successfully used in different scenarios~\cite{Lorena2018}. Most complexity measures are computationally expensive; one approach to reduce the associate computational cost is to train a metaregressor using traditional meta-features as input~\cite{garcia2020simulated}. The metaregressor estimates complexity values for any dataset at low cost; the predicted values are called simulated meta-features.

Meta-features can be transformed to summarize the data, e.g., by reducing data dimensionality. Principal Component Analysis PCA~\citep{Hotelling1933} is the most straightforward and generic approach, even though it ignores the metatarget. For example, after running PCA, \citet{Bilalli2017} select the most relevant components (according to the cumulative total variance); next, a filter capturing the correlation with the metatarget identifies the most discriminating variables.
\citet{Munoz2018} propose a new dimensionality reduction technique suited for data visualization; the most statistically significant traditional meta-features representing the hardness of the datasets are first identified, after which the metadataset is transformed into a 2-dimensional space to search for an optimal projection through an iterative optimization process; the new 2D space can project different model-performance footprints to investigate their strengths and weaknesses.


A different approach that has achieved popularity in recent years invokes Deep Neural Networks (DNNs). A strength behind DNNs is the capacity to learn data characteristics from a diverse and large amount of data. DNNs have had a strong impact in application areas such as speech recognition and image understanding~\cite{Deng:2014}. However, its use in meta-learning is still incipient and requires further investigation. A few studies explore deep learning for feature generation, representing different tasks and datasets in terms of embeddings generated by pre-trained DNNs~\cite{Achille2019}. \citet{Gosztolya2017} solve different automatic speech recognition tasks through a two-step learning process: performing classification with DNNs, followed by the extraction of intrinsic features from the DNN output. In the second phase, features were used to improve model predictions. The same strategy is explored by~\citet{Notley2018}, using DNNs to extract features from both images and numeric data. 


Our hypothesis is that DNNs provide the means to extract \textit{intrinsic} features from data. Our approach lies between transformation and deep learning methods since a DNN is induced from traditional meta-features and the knowledge captured by the hidden layers are used to generate abstract meta-features.
In the process, traditional meta-features are transformed into latent variables used by the deep learning model to make predictions. 
Once extracted, abstract meta-features can be used by any meta-learning algorithm. 

Our approach differs from traditional meta-learning settings. While model-based and landmarking induce a learning model for each dataset (metainstances) and extract features from this model, in the current approach, a model is induced using the metabase (all datasets). This is akin to the simulated meta-feature approach; however, instead of using the model's predicted values, we use the abstract representation of the DNN model to extract meta-features. 


\section{Problem Statement}
\label{sec:back}

Given a classification task on a dataset $\mathcal{D}$ with $n$ instances, our goal is to compute a meta-feature $f$ on $\mathcal{D}$. A meta-feature is usually a hand-crafted characterization function capturing a specific property of interest on a given task. Meta-features are regarded as a form of meta-knowledge collected over a distribution of tasks to learn \emph{how to learn}. Not all meta-features are informative, and some of them are very task specific. Learning relevant meta-features can prove useful in identifying hidden relationships across tasks, and is necessary to build accurate meta-learners. 

Knowledge extracted across tasks, a.k.a. meta-knowledge, is key to the success of meta-learning by obviating learning from scratch on new tasks. By exploiting meta-knowledge, the metalearner can effectively construct an optimal solution based on past experience \cite{hospedales2020metalearning, vanschoren2018meta}. For example, a meta-learner can identify that a new task is similar to previous tasks and warm-start a similar model with near optimal hyperparameters. This avoids the --sometimes painstakingly-- slow processes of error and trial in building a new model. Meta-knowledge can be understood as meta-features, model hyperparameters, performance measures, etc. In our work, meta-knowledge consists of meta-features and performance measures gathered from previous tasks.

We formally define the process of meta-feature extraction as a function $f$ that receives as input a dataset $\mathcal{D}$, and returns as output a set of $k$ values characterizing the dataset \cite{rivolli2018towards}:

\begin{equation}
    f(\mathcal{D})=\sigma(m(\mathcal{D})),
\end{equation}

\noindent
where $m$ is a characterization metric mapping  $\mathcal{D}\rightarrow\mathbb{R}^{k'}$, $k'$ is the original number of meta-features, and $\sigma$ is a summarization function mapping $\mathbb{R}^{k'} \rightarrow \mathbb{R}^k$. The purpose of the summarization function is to reduce the size of the output to a fixed size. A dataset $\mathcal{D}$ is characterized by a meta-feature space $\mathcal{F}$. Our goal is to find a subset of meta-features, $F \subset \mathcal{F}$, capturing relevant task information. 

\section{Abstract Meta-features}
\label{sec:abs_mfs}
\begin{figure}[t!]
        \centering
        \includegraphics[width=0.45\textwidth]{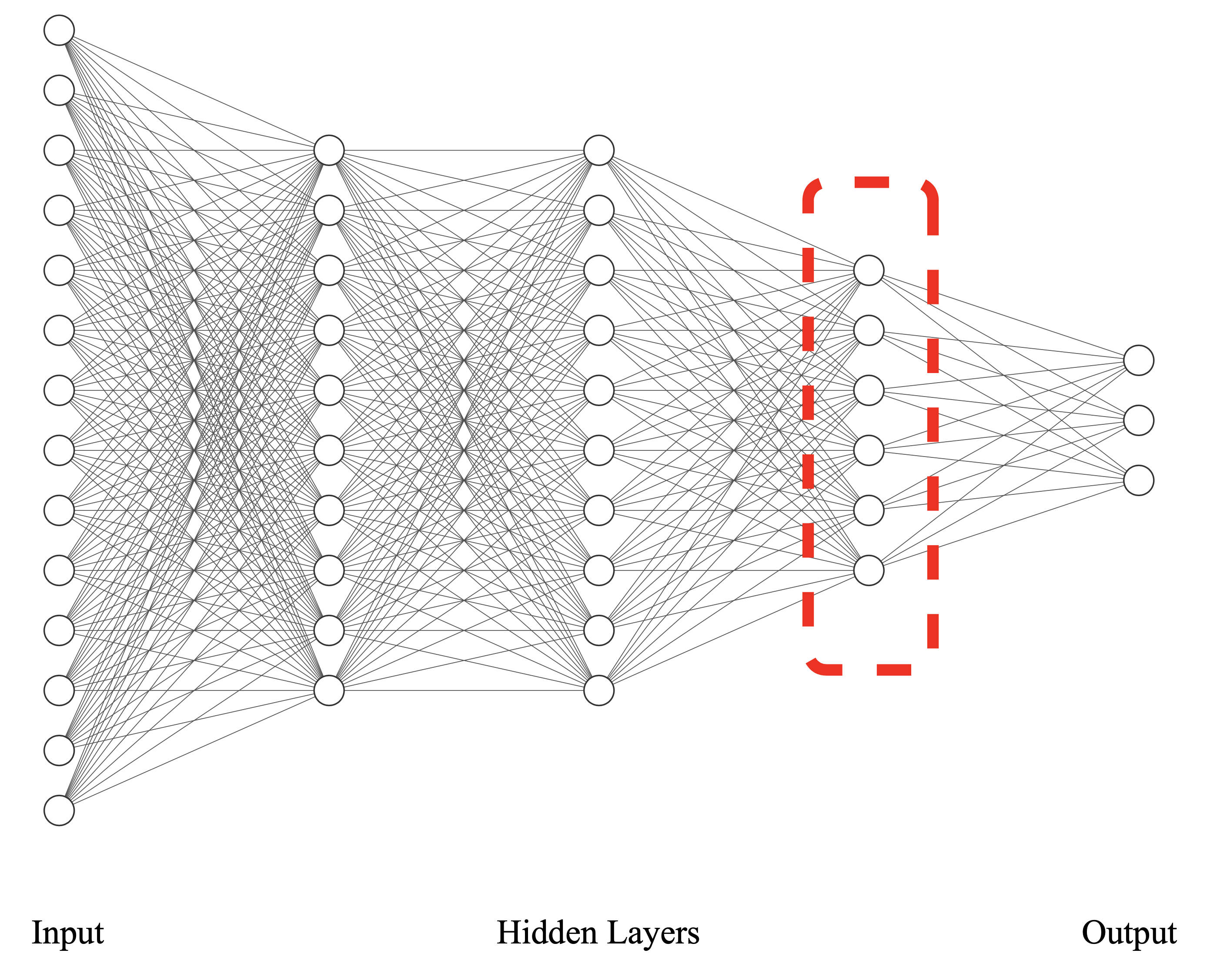}
        \caption{A deep neural network for learning abstract meta-features. The last hidden layer is used to extract the learnt latent variables $\{z_i\}$ to produce abstract meta-features.}
        \label{fig:nn}
\end{figure}

We propose a novel approach to learning new abstract meta-features by constructing new representations from traditional meta-features using a deep neural network. A neural network --parameterized by $\mathbf{w}$-- is a universal function estimator. The goal is to learn a function $g$ to predict the true target $y$ through an approximation $\hat{y}$, where $g(\mathbf{x})=\hat{y}$. Training is achieved by computing gradients of the loss function with respect to the weights $\nabla_\mathbf{w} J(\mathbf{w})$ and then back-propagating the errors through the network to update the weights. 
The architecture of a neural network is described by the number of neurons per layer and by the number of hidden layers $\ell_{h}^{n}$, where $h$ is the index for the hidden layer and $n$ is the total number of neurons in that layer. Forward propagating input $\mathbf{x}$ through the network leads to a sequence of non-linear transformations; non-linearity is achieved via an activation function $\phi$. Increasing the number of hidden layers and neurons allows the neural network to approximate highly non-linear functions. Each layer contains a learnt latent representation of the input data. The last hidden layer comprises the final learnt latent variables, $\{z_i\}$, where each latent variable $z_i$ is a representation of the original input in an abstract space. The number of latent variables is user-defined by controlling the number of neurons in that layer. By training a deep neural network on the traditional meta-feature space $\mathcal{F}_t$, we can learn a new latent representation $\mathcal{F}_a$ (abstract meta-features). The resulting deep neural network serves as a feature extractor where the learnt latent space $Z$ is extracted from the last hidden layer. This process is highlighted in Figure~\ref{fig:nn}.

\begin{table}[h!]
\centering
\small
{%
\begin{tabular}{@{}ll@{}}
\toprule
\textbf{Hyper-parameter} & \textbf{Value} \\ \midrule
Learning rate            & 0.005          \\
Hidden layers            & 5              \\
Latent variables         & 16             \\
Criterion                & $\textnormal{Smooth}_{L_{1}}$   \\
Optimizer                & Adam           \\
Activation               & ReLU           \\ \bottomrule
\end{tabular}%
}
\caption{AbstractNet hyper-parameters.}
\label{tab:hyp}
\end{table}

\subsection{AbstractNet}
Our methodology constructs a deep neural network (DNN) to act as a feature extractor on each pair of input and output meta-instances. After providing traditional meta-features as input and a performance measure of three different algorithms as target, we train our \texttt{AbstractNet} to learn a meaningful abstract representation of the meta-dataset. \texttt{AbstractNet} consists of $5$ fully connected layers with sixty four neurons in each of the first four layers ($\ell^{64}_{h}, 1 \le h \le 4$), and a final latent layer with sixteen neurons ($\ell^{16}_5$). Our target is a three dimensional output consisting of a performance estimation for each of three learning algorithms applied to a given task (represented using meta-features). Non-linearity between layers is achieved via the ReLU activation function,

\begin{equation}
    \phi(q)=\max(0, q),
\end{equation}

\noindent
where $q=\mathbf{wx}+\mathbf{b}$ is the linear transformation of the input. Variance is controlled via dropout; regularization is applied between layers two and four with probabilities $p=(0.1, 0.05)$ respectively. We selected the smooth $L_1$ loss function \cite{girshick2015fast} as our criterion function:

\begin{equation}
    J(\hat{y}, y) = \frac{1}{n} \sum_i \nu(\hat{y}_i-y_i),
\end{equation}

\noindent
where the summation goes over all training examples, $\hat{y}$ and $y$ are the estimated and true response values respectively, and $\nu$ is defined as 

\begin{equation}
    \nu(u) =
        \left\{ \begin{array}{ll}
            (0.5*u^2) / \lambda   &\textnormal{if}\quad|u| < \lambda \\
            |u|-0.5*\lambda       &\textnormal{otherwise.}
        \end{array} \right.
\end{equation}

\noindent
Parameter $\lambda$ specifies the threshold that defines the step function. The smooth $L_1$ loss is less sensitive to outliers, has a smoother landscape, and prevents exploding gradients. The list of full neural network hyper-parameters is available in Table \ref{tab:hyp}. Once the deep neural network is trained, we forward propagate the meta-dataset validation partition to extract the latent variables $\{z_i\}$ from the last hidden layer to induce our meta-models.


\section{Experiments}
\label{sec:methods}

\begin{figure*}[ht!]
    \centering
    \begin{tabular}{cc}
       {\includegraphics 
       [width = \textwidth]
       {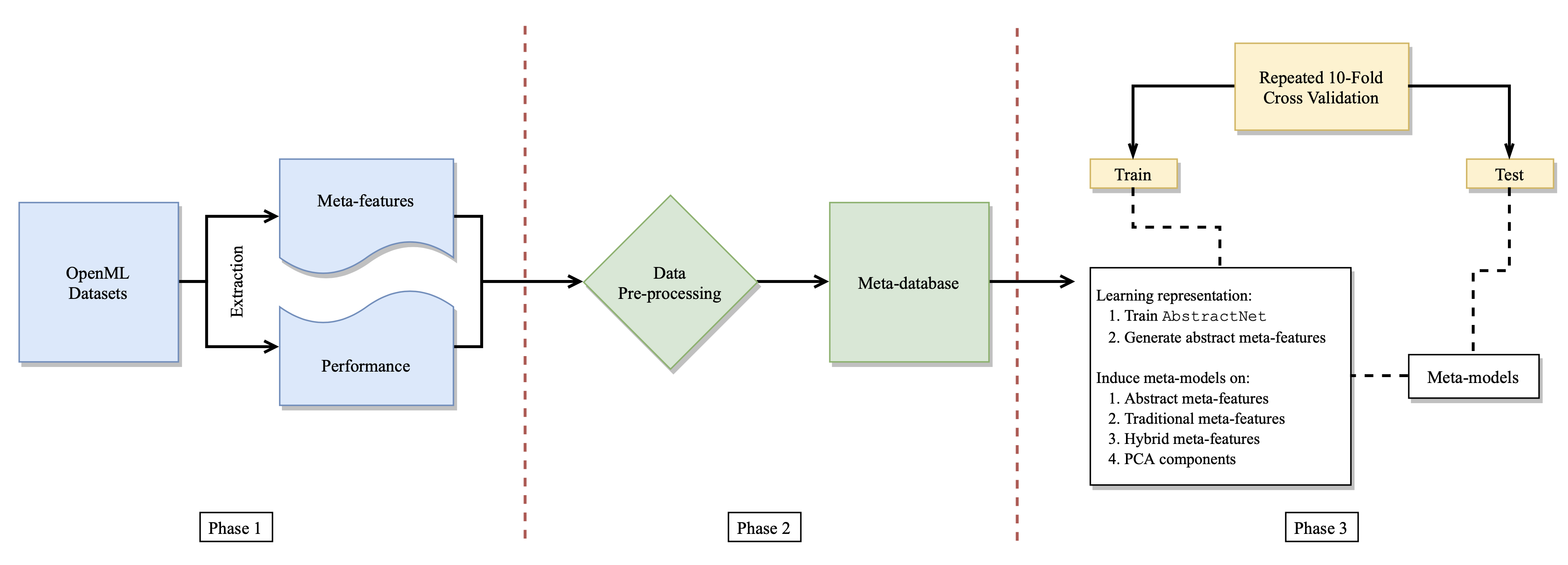}}
    \end{tabular}
    \caption{Sequence of steps to generate abstract meta-features and to assess meta-model performance.}
    \label{fig:exp_methodology}
\end{figure*}

In this section we describe the experimental design to induce abstract meta-features. Figure~\ref{fig:exp_methodology} provides an overview of the entire process, with three different phases corresponding to data characterization, meta-database construction, and induced meta-model evaluation. We explain these steps in detail next.


\subsection{Datasets and Performance Evaluation}


Phase 1 of our experiments includes dataset selection, performance evaluation, and meta-feature extraction. We collected a total of $517$ classification datasets from OpenML~\cite{Vanschoren:2014}, a free scientific platform for standardization of experiments and sharing of empirical results. These datasets were selected following some criteria: the number of features does not exceed $500$; there are no missing values; there are at least $2$ classes; and the minority class must have at least $10$ examples. 

Next, we evaluated three common learning algorithms: Support Vector Machine (SVM), Random Forest (RF), and Multilayer Perceptron (MLP) on each dataset, recording their generalization performance in terms of the Area Under the ROC Curve (AUC)~\citep{Hand2001}.  

In parallel, we extracted the traditional family of meta-features using the \texttt{PyMFE} library~\citep{alcobacca2020mfe} for these categories: general, statistical, info-theoretical, concept, model-based and landmarking.  The functions \emph{max, min} and \emph{mean} were used to summarize multi-valued results.

\subsection{Meta-database}
In phase 2, we combined meta-features and performance values per dataset to construct a meta-database, $\mathcal{D}_{meta}$. A pre-processing step was also performed at the meta-database by removing meta-instances with more than $100$ missing values, meta-features with more than $70\%$ of missing values, constant meta-features, and highly correlated meta-features. The remaining missing values were imputed using a $k$-NN with $k=10$. Lastly, we generated a meta-database with three different targets, one for each performance prediction problem (SVM, RF, MLP). The final meta-database has a total of $517$ meta-examples (instances) and $265$ meta-features (characteristics); it was used to train the deep neural network to induce the latent abstract meta-features. 


\subsection{Evaluation}


In phase 3, we evaluated our approach under four different settings: 
\begin{itemize}
    \item  \textbf{Abstract}: this strategy explores our approach alone, by constructing abstract meta-features through the \texttt{AbstractNet} and extracting a latent representation from the last hidden layer. 

    \item \textbf{Traditional}: we induce meta-models on traditional meta-features only.
    
    \item \textbf{Hybrid}: we induce meta-models using a combination of traditional and abstract meta-features.

    \item \textbf{PCA}: as a baseline for comparison, PCA~\citep{Hotelling1933} is invoked to transform traditional meta-features through linear combinations. PCA also generates latent features keeping the components with 95\% of the cumulative variance.

\end{itemize}

\noindent The hybrid approach is instrumental to assess feature relevance; it can show the value of abstract meta-features in predicting model performance over traditional meta-features. 

We performed $10 \times 10$-fold cross-validation; in each iteration, nine folds are used to obtain the abstract meta-features (\texttt{AbstractNet}) and induce the different meta-models; and the remaining fold is used to validate the meta-models. We employed three learning algorithms: Decision Trees, Random Forest, and Support Vector Machines as meta-inducers. Finally, we evaluated the performance of our models on the validation sets and report the Root Mean Squared Error (RMSE) and the coefficient of determination ($R^2$). We also repeated the experiments 10 times with different seeds to perform statistical validations using the Hierarchical Bayesian correlated t-test~\citep{Benavoli2017}. 
Here, we compared the performance of the meta-inducers using distinct subsets of meta-features. 
The test evaluates in pairs, resulting in probabilities concerning which approach is better (left and right) for a particular evaluation measure. It also defines a region of equivalence (rope) that indicates the probability that the difference in performance is insignificant.
The complete experimental methodology is shown in Table~\ref{tab:exp_setup}.




\begin{table}[t!]
    \centering
    \small
    \begin{tabular}{lll}
            
            \toprule
            \textbf{Element} & \textbf{Feature} & \textbf{Value} \\ 
    
            \midrule
            \multirow{3}{*}{Base level} & Datasets & 517 \\
             & Target Algorithms & SVM, RF, MLP \\
             & Performance & AUC \\
            
            \midrule
            \multirow{4}{*}{Meta-features} & Abstract & 16 \\
            & Traditional & 154 \\
            & Hybrid & 170 \\
            & PCA & 60 (95\% of var) \\
            
            \midrule
            
            \multirow{4}{*}{Meta-level} & Tasks & 3 (base level) \\
            & Regressor & RF, DT, SVM \\
            & Resampling & 10 x 10-CV \\
            & Performance & RMSE, $R^2$ \\
        
            \midrule
            
            Statistical Validation & Test & Hier. Bayesian \\
          
            \bottomrule
        \end{tabular}
        \caption{Experimental settings across base and meta-levels.}        \label{tab:exp_setup}
\end{table}

\section{Results and Discussion}
\label{sec:results}

Experimental results are shown in Table~\ref{tbl:res}. Results in bold stand for best results. As we can see, abstract meta-features achieve best RMSE and $R^2$ scores across all four settings. For instance, abstract meta-features perform vastly better than traditional meta-features with Decision Trees; the dimensionality of abstract meta-features is sixteen, while that of traditional meta-features is one hundred and fifty four. The reduced size of the space of abstract meta-features leads to high-generalization meta-models. Similar results are seen when different learning algorithms are used to induce meta-models. PCA transformation ranked lowest in terms of performance across the four settings; we hypothesize that PCA's inherent linear components are not expressive enough. \texttt{AbstractNet} is capable of learning highly abstract representations to capture complex relationships between meta-features and the target variable. 

\begin{figure}[!hbt]
        \centering
        \includegraphics[width=\columnwidth]{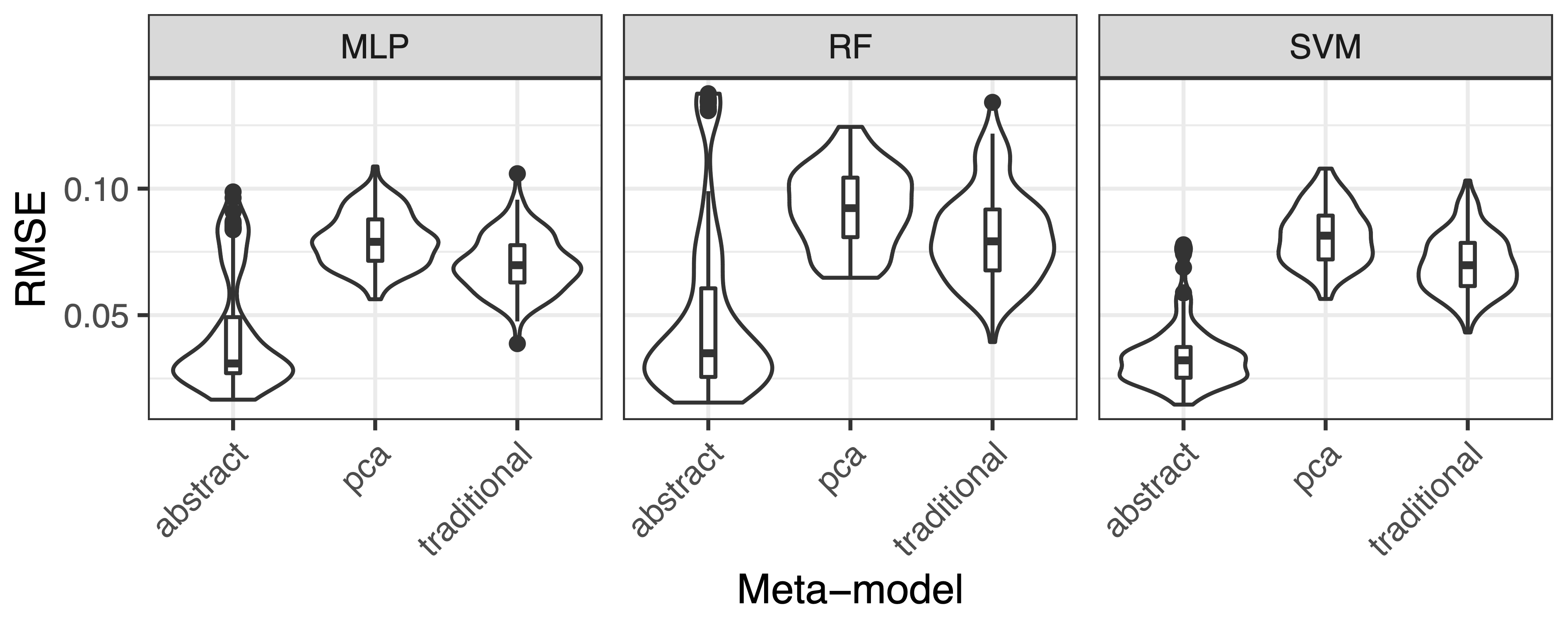}
        \caption{Violin plot of RMSE scores of meta-models on three approaches: abstract, traditional, and PCA.}
        \label{fig:rmse_plot}
\vspace{1.5cm}
        \centering
        \includegraphics[width=\columnwidth]{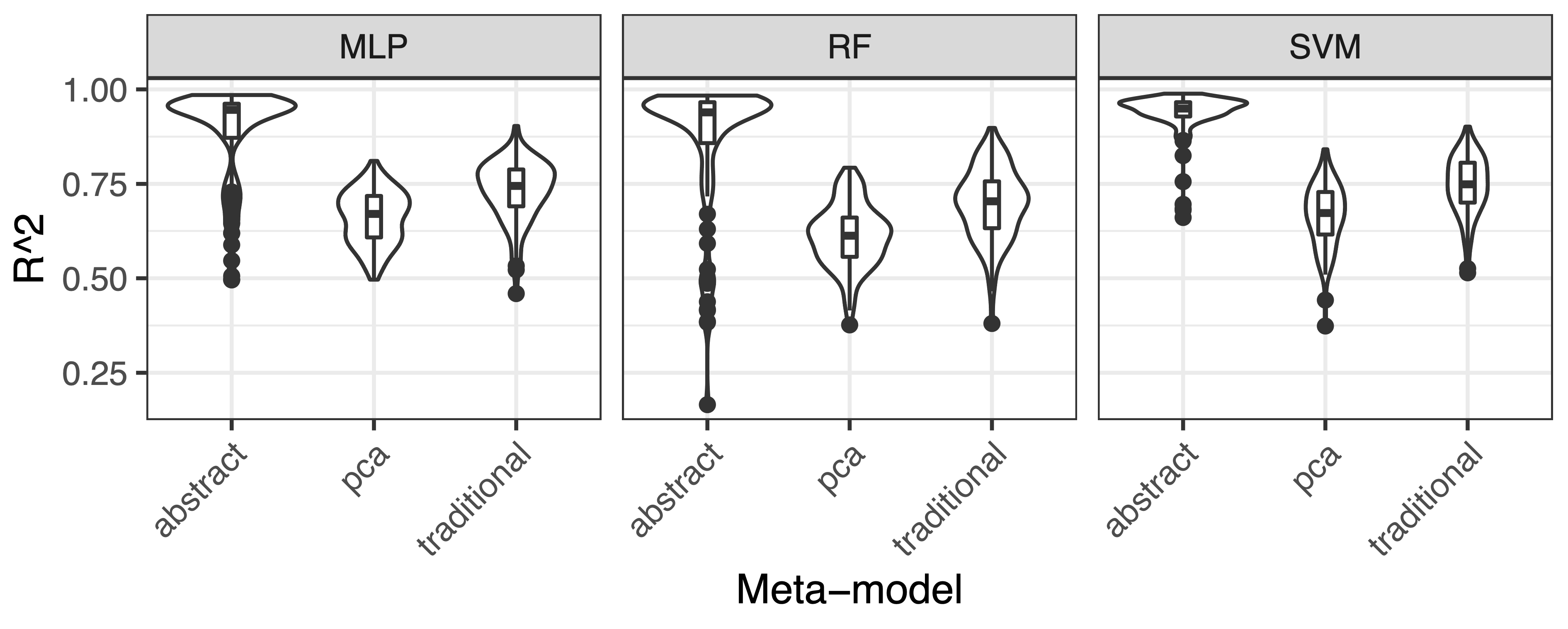}
        \caption{Violin plot of $R^2$ scores of meta-models on three approaches: abstract, traditional, and PCA.}
        \label{fig:r2_plot}
\end{figure}

We can see that abstract meta-features closely follow the true AUC values across datasets, while traditional and PCA meta-features exhibit instability and poor performance. Variance along our performance metrics per learning algorithm is shown in Figures~\ref{fig:rmse_plot}, \ref{fig:r2_plot}. The hybrid approach allows us to combine traditional and abstract meta-features. By ranking the top fifteen most important features using Gini index from the Random Forest meta-model, seven out of sixteen learnt abstract meta-features ranked in the top fifteen features.
 
Table~\ref{tbl:stats} shows probabilities obtained with the Bayesian Hierarchical $t$-test over different meta-databases and performance values. Abstract meta-features improved the use of traditional and PCA meta-features significantly, confirming the generalization ability of our approach. Unlike PCA, our deep neural network learnt a non-linear abstract transformation of traditional meta-features while increasing their predictive power.

\begin{table*}[t!]
\small
\resizebox{\textwidth}{!}{%
\begin{tabular}{lllllllll}
\toprule
\multirow{2}{*}{\textbf{Inducer}} & \multirow{2}{*}{\textbf{Meta-features}}  & \multicolumn{3}{c}{\textbf{$R^2$ Score}}                                                   &  & \multicolumn{3}{c}{\textbf{RMSE}}                                                 \\ 
             &  & SVM AUC                & RF AUC                 & MLP AUC                &  & SVM AUC                & RF AUC                 & MLP AUC                \\  \cline{3-5} \cline{7-9} \addlinespace
DT           & Abstract      & \textbf{0.906 (0.072)} & \textbf{0.839 (0.17)}  & \textbf{0.864 (0.129)} &  & \textbf{0.043 (0.014)} & \textbf{0.057 (0.032)} & \textbf{0.048 (0.022)} \\
             & Traditional   & 0.554 (0.132)          & 0.510 (0.155)          & 0.561 (0.14)           &  & 0.101 (0.019)          & 0.114 (0.024)          & 0.097 (0.018)          \\
             & Hybrid        & 0.877 (0.078)          & 0.803 (0.163)          & 0.841 (0.128)          &  & 0.050 (0.015)          & 0.067 (0.031)          & 0.054 (0.022)          \\
             & PCA           & 0.424 (0.136)          & 0.384 (0.132)          & 0.418 (0.12)           &  & 0.121 (0.018)          & 0.131 (0.022)          & 0.117 (0.016)          \\ \cline{3-5} \cline{7-9} \addlinespace
             
RF           & Abstract      & 0.925 (0.059)          & \textbf{0.856 (0.17)}  & \textbf{0.882 (0.125)} &  & 0.038 (0.012)          & \textbf{0.052 (0.033)} & \textbf{0.044 (0.023)} \\
             & Traditional   & 0.761 (0.076)          & 0.715 (0.094)          & 0.752 (0.079)          &  & 0.071 (0.012)          & 0.081 (0.018)          & 0.070 (0.011)          \\
             & Hybrid        & \textbf{0.928 (0.055)} & 0.851 (0.167)          & 0.880 (0.123)          &  & \textbf{0.037 (0.012)} & 0.054 (0.032)          & 0.044 (0.023)          \\
             & PCA           & 0.688 (0.083)          & 0.641 (0.085)          & 0.685 (0.075)          &  & 0.081 (0.012)          & 0.092 (0.016)          & 0.080 (0.011)          \\ \cline{3-5} \cline{7-9} \addlinespace
             
SVM          & Abstract      & \textbf{0.889 (0.050)} & \textbf{0.862 (0.109)} & \textbf{0.873 (0.083)} &  & \textbf{0.062 (0.008)} & \textbf{0.076 (0.017)} & \textbf{0.068 (0.011)} \\
             & Hybrid        & 0.829 (0.049)          & 0.774 (0.100)          & 0.805 (0.073)          &  & 0.070 (0.007)          & 0.083 (0.016)          & 0.074 (0.010)          \\
             & Traditional   & 0.685 (0.082)          & 0.603 (0.108)          & 0.644 (0.081)          &  & 0.088 (0.010)          & 0.102 (0.016)          & 0.092 (0.009)          \\
             & PCA           & 0.715 (0.067)          & 0.640 (0.105)          & 0.704 (0.075)          &  & 0.082 (0.009)          & 0.096 (0.016)          & 0.085 (0.009) \\
             \bottomrule
\end{tabular}}
\caption{Generalization performance of meta-models using different meta-databases containing all families of meta-features.}
\label{tbl:res}
\end{table*}


\begin{table}[]
\centering
\small
\begin{tabular}{lllll}
\toprule
\textbf{Meta-databases}         & \textbf{Measure} & \textbf{left}           & \textbf{rope}  & \textbf{right}          \\ \midrule
\multirow{2}{*}{Traditional $\times$ Abstract} & RMSE    & 0.001          & 0.000 & \textbf{0.999} \\
                       & $R^2$ Score      & 0.000          & 0.000     & \textbf{1.000} \\ \addlinespace \midrule
\multirow{2}{*}{PCA $\times$ Abstract}         & RMSE    & 0.000          & 0.000     & \textbf{1.000} \\
                       & $R^2$ Score      & 0.001          & 0.000     & \textbf{0.999} \\ \addlinespace \midrule
\multirow{2}{*}{Traditional $\times$ PCA}      & RMSE    & 0.292 & \textbf{0.696} & 0.012           \\
                       & $R^2$ Score      & \textbf{0.911} & 0.000     & 0.089    \\
\bottomrule    
\end{tabular}
\caption{Hierarchical Bayesian statistical probabilities by comparing pairs of meta-databases. The columns \emph{left} and \emph{right} indicate each meta-database's probability outperforming the other. The \emph{rope} column indicates the probability of them being similar.}
\label{tbl:stats}
\end{table}



\section{Conclusions}
\label{sec:conclusion}


Data-driven meta-learning (MtL) requires new forms of data characterization. Given the difficulty inherent to the size of the meta-feature space, this paper explores a promising direction to solve the meta-feature selection problem by learning abstract meta-features via a deep neural network. We introduce and discuss data characterization as meta-knowledge. In order to optimally meta-learn over a distribution of tasks, the right form of meta-knowledge is required \emph{a priori}. By defining our meta-objective as generalization performance, we construct a deep neural network to learn an abstract representation of traditional meta-features, i.e., we generate abstract meta-features as meta-knowledge to be used by our meta-models. We contend that abstract meta-features are more expressive and effectively capture hidden variable relationships. 

Our experimental results demonstrate the efficacy of abstract meta-features as strong predictors of generalization performance, while reducing the size of the meta-feature space. Feature importance values computed on hybrid meta-features show that abstract meta-features frequently achieve top results. Our results show that PCA linear transformations are not as expressive as the non-linear transformations learnt by our deep neural network. 

\subsection{Limitations \& Future work}
There is no clear subset of traditional meta-features capable of capturing task properties well over highly diversified tasks. Just as deep learning usually outperforms handcrafted features, learning meta-features with DNN can offer great insight into improving the data characterization process. Future research directions involve exploring abstract meta-features in a cost-effective fashion, and decomposing complex task characteristics as functions of simpler building properties. The goal is to identify fundamental characteristics that cover a broad spectrum of complex tasks. 




\vspace*{10mm}



\bibliography{references}

\end{document}